# CKSP: Cross-species Knowledge Sharing and Preserving for Universal Animal Activity Recognition

Axiu Mao [a], Meilu Zhu [b*], Zhaojin Guo [c], Zheng He [c], Tomas Norton [d], Kai Liu [c*]

[a] School of Communication Engineering, Hangzhou Dianzi University, Hangzhou, 310018, Zhejiang Province, China. axiu.mao@hdu.edu.cn (Axiu Mao)

[b] Department of Mechanical Engineering, City University of Hong Kong, Kowloon Tong, 999077, Hong Kong SAR, China. meiluzhu2-c@my.cityu.edu.hk (Meilu Zhu)

[c] Department of Infectious Diseases and Public Health, Jockey Club College of Veterinary Medicine and Life Sciences, City University of Hong Kong, Kowloon Tong, 999077, Hong Kong SAR, China. zggguo2-c@my.cityu.edu.hk (Zhaojin Guo); zhenghe8-c@my.cityu.edu.hk (Zheng He); kailiu@cityu.edu.hk (Kai Liu)

[d] Department of Biosystems, Division Animal and Human Health Engineering, M3-BIORES, Katholieke Universiteit van Leuven, Kasteelpark Arenberg 30, B-3001, Leuven, Belgium. tomas.norton@kuleuven.be (Tomas Norton)

* Corresponding author. Kai Liu: kailiu@cityu.edu.hk;

Meilu Zhu: meiluzhu2-c@my.cityu.edu.hk.

**Abstract**

Deep learning techniques are dominating automated animal activity recognition (AAR) tasks with wearable sensors due to their high performance on large-scale labelled data. However, current deep learning-based AAR models are trained solely on datasets of individual animal species, constraining their applicability in practice and performing poorly when training data are limited. In this study, we propose a one-for-many framework, dubbed Cross-species Knowledge Sharing and Preserving (CKSP), based on sensor data of diverse animal species. Given the coexistence of generic and species-specific behavioural patterns among different species, we design a Shared-Preserved Convolution (SPConv) module. This module assigns an individual low-rank convolutional layer to each species for extracting species-specific features and employs a shared full-rank convolutional layer to learn generic features, enabling the CKSP framework to learn inter-species complementarity and alleviating data limitations via increasing data diversity. Considering the training conflict arising from discrepancies in data distributions among species, we devise a Species-specific Batch Normalization (SBN) module, that involves multiple BN layers to separately fit the distributions of different species. To validate CKSP's effectiveness, experiments are performed on three public datasets from horses, sheep, and cattle, respectively. The results show that our approach remarkably boosts the classification performance compared to the baseline method (one-for-one framework) solely trained on individual-species data, with increments of 6.04%, 2.06%, and 3.66% in accuracy, and 10.33%, 3.67%, and 7.90% in F1-score for the horse, sheep, and cattle datasets, respectively. This proves the promising capabilities of our method in leveraging multi-species data to augment classification performance.

**Nomenclature**

*Symbols*

| | |
|---|---|
| $s$ | Animal species |
| $x^s$ | Feature maps of species $s$ at a given layer |
| $x_o^s$ | Output of $x^s$ fed into a low-rank convolution layer of species $s$ |
| $W_{LR}$ | The original parameter tensor in the low-rank convolution layer |
| $B$ | A matrix that is initialised as zero |
| $A$ | A matrix that is initialised with random Gaussian values $\mathcal{N}(0, \sigma^2)$ |
| $\gamma^s$ | A moving mean in the batch normalization layer of species $s$ |
| $\beta^s$ | A moving variance in the batch normalization layer of species $s$ |
| $x_b^s$ | Feature map of the $b$-th sample within a batch of species $s$ |
| $y_b^s$ | Output of $x_b^s$ fed into a BN layer of species $s$ |
| $\mathcal{L}^s$ | Class-balanced focal loss of species $s$ |
| $\mathcal{L}$ | Overall loss function |

*Abbreviations*

| | |
|---|---|
| AAR | Animal activity recognition |
| CKSP | Cross-species Knowledge Sharing and Preserving |
| SPConv | Shared-Preserved Convolution |
| SBN | Species-specific Batch Normalization |
| MLP | Multilayer perceptron |
| CNN | Convolutional neural network |
| RNN | Recurrent neural network |
| LoRA | Low-Rank Adaptation |
| LRConv | Low-Rank Convolution |
| FRConv | Full-rank convolution |

# 1. Introduction

Automated animal activity recognition (AAR) with wearable sensors empowers caretakers to continuously and remotely monitor behavioural variations in animals, considerably decreasing workloads and expenses in veterinary practices while enhancing the efficiency and sustainability of livestock management (Mao et al., 2023a). Wearable sensors are often incorporated into both research-oriented and commercial devices for specific applications, like the Whistle Fit (Chambers et al., 2021) and Ceres Tags (Wang et al., 2023). These devices are attached to various animal body parts, including necks, ears, and legs, to capture motion data like acceleration and angular velocity. These data are then processed and analysed using smart computing techniques to achieve accurate classification of animal behaviours like cattle grazing and walking (Arablouei et al., 2023a), and sheep scratching and resting (Kleanthous et al., 2022a).

Currently, deep learning is dominating wearable sensor-aided AAR tasks owing to their exceptional feature extraction abilities, showcasing favourable performance in discriminating animal behaviours across a wide range of scenarios (Kleanthous et al., 2022b; Riaboff et al., 2022). Arablouei et al. (2021) examined the application of multilayer perceptron (MLP) in cattle behaviour recognition, achieving a higher accuracy of 93.4% than several machine learning methods such as support vector machines. Their developed MLP model was subsequently utilised in further research, consistently exhibiting promising results (Arablouei et al., 2023a, 2023b). Convolutional neural networks (CNNs), as the most commonly applied method in AAR tasks, have achieved high accuracies often exceeding 90%, attributed primarily to their capabilities to capture local temporal dependencies and exhibit scale invariance (Mao et al., 2023). Furthermore, recent studies have explored combining CNNs with recurrent neural networks (RNNs) for classifying animal behaviours using sensor data, with the hybrid models tending to exhibit desirable performance than pure CNN- and RNN-based models (Liseune et al., 2021; Wang et al., 2023).

Despite the satisfactory performance, current deep learning-based AAR methods (Arablouei et al., 2023a; Riaboff et al., 2022; Wang et al., 2023) still have the following issues. (1) These approaches are generally trained on datasets of individual animal species. This greatly constrains their applicability in real-world scenarios, as they cannot be directly applied to different species due to data discrepancy (e.g., different sampling rates over identical time spans or distinct movement patterns). Meanwhile, designing and training unique models for each species is time-consuming and also leads to inconvenience for deployment. (2) The exceptional performance of deep learning-based methods normally hinges on the availability of large-scale, labelled training data. However, the availability of datasets for some species is sometimes constrained by some inevitable factors, such as the laborious and time-consuming labelling process, insufficient animal objects, collection difficulty, and so on. Insufficient training data results in weak feature representation ability, inferior generalization performance, or even optimization failure (C. Li et al., 2021). Therefore, existing one-for-one models are obviously not promising solutions for the AAR task.

An intuitive idea to avoid the above issues is developing a universal AAR framework that is applicable to various animal species. However, to guarantee that this one-for-many framework can outperform existing one-for-one models, we need to achieve two goals: seeking common ground and reserving differences between behaviours of different species. Firstly, some behaviours of different species demonstrate similarities in motion patterns, e.g., grazing for sheep and cattle and standing for all species (Arnold, 1984; Patkowski et al., 2019; Pluta et al., 2013). When the training samples of one class for a species are limited or the diversity of training data for a species is poor, constructing multi-species training datasets can improve data quality by sharing similar behaviour patterns (common ground) among various species, thereby boosting the performance of a universal AAR framework. Secondly, inherent differences in behaviours between different species, primarily manifested in distinct movement patterns and divergent feature distributions, inevitably pose challenges to the stable training of a universal framework. Inter-species variation in motion patterns affects a universal AAR framework in capturing invariant features across different species while hindering model convergence. Distinct distributions among species also make it challenging to learn universal global statistical measures, which contradicts the assertion that neural networks' high performance typically depends on a well-normalised data distribution (Wang et al., 2019). Hence, exploring a practical solution that simultaneously exploits cross-species behavioural similarities and preserves their unique characteristics becomes imperative.

To achieve the afore-mentioned objectives, we attempt to establish a universal AAR framework,

dubbed Cross-species Knowledge Sharing and Preserving (CKSP), based on sensor data across diverse animal species. Essentially, the CKSP is versatile, being applicable to different species with distinct interested behaviours. To capture similarities and preserve discrepancies of behaviours between different species, a Shared-Preserved Convolution (SPConv) module and a Species-specific Batch Normalization (SBN) module are designed in the feature extraction process. The SPConv module assigns individual low-rank convolutional layers to each species for extracting species-specific features while employing a shared full-rank convolutional layer to learn shared generic features. The SBN module involves multiple BN layers that separately fit the distributions of different species. To demonstrate the effectiveness of our method, the proposed CKSP is trained concurrently on three publicly available datasets sourced from horses (Kamminga et al., 2019a), sheep (Kleanthous et al., 2022), and cattle (C. Li et al., 2021). Its classification performance is compared against that of the baseline model (Single-Net) solely trained on individual species data. To be summarized, our contributions are as follows.

- The proposed CKSP leverages multi-species sensor data to construct a universal AAR framework. This framework is adaptable to various species with distinct interested behaviours, implicitly addressing data limitation issues encountered in single-species model training. To the best of our knowledge, we are the pioneers in developing a universal AAR framework applicable to diverse species, meanwhile opening up a new perspective in addressing data limitation challenges.

- We devise an SPConv module, which assigns individual low-rank convolutional layers to each species for extracting species-specific features, in addition to a shared full-rank convolutional layer for learning generic features. This way well considers the coexistence of shared generic features and species-specific movement characteristics among different species, effectively alleviating the dilemma of feature learning encountered when relying solely on a unified shared feature extraction mechanism.

- Given that optimal neural network performance requires a well-normalised data distribution, yet different species exhibit distinct data distributions, we devise an SBN module. This module allocates a separate BN layer to each species, allowing independent adaptation to their unique distribution characteristics, and thus enhancing normalization efficacy across diverse species.

- Experiments performed on public datasets from three species (Kamminga et al., 2019a; Kleanthous et al., 2022a; C. Li et al., 2021) showcase that our CKSP approach significantly enhances classification performance over the Single-Net trained exclusively on single-species data.

## 2. Proposed method

The proposed Cross-species Knowledge Sharing and Preserving (CKSP) method aims to develop a universal AAR framework based on multi-species sensor datasets, and such a framework is broadly applicable to different species while tackling the data limitation challenge typically encountered when relying solely on individual species data. Figure 1 illustrates the overall workflow of the CKSP framework, mainly encompassing three parts, i.e., data preprocessing, feature extraction, and behavioural classification.

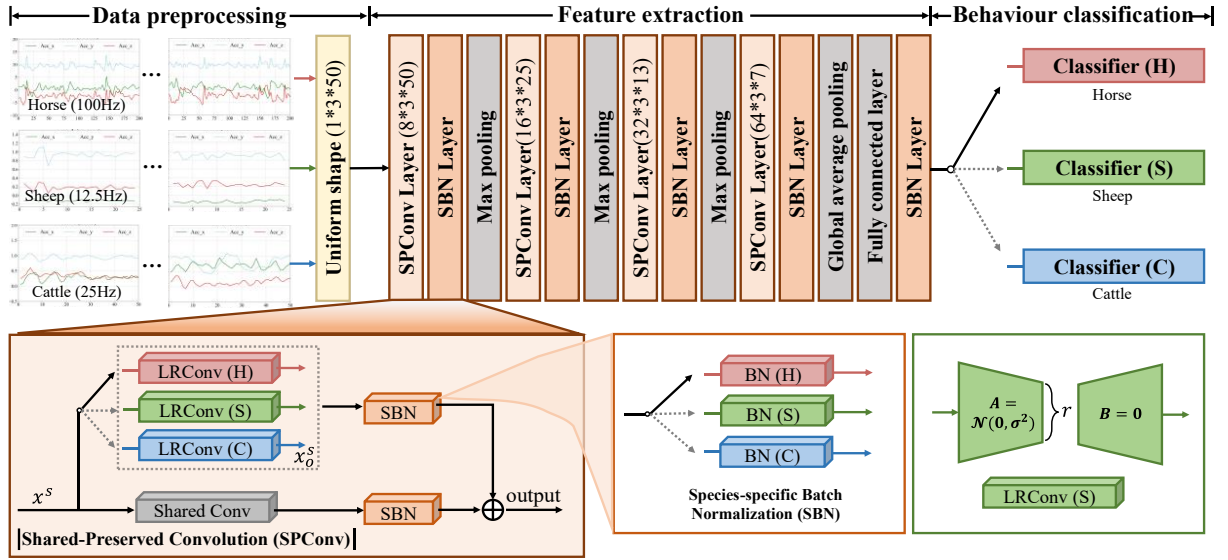

**Fig. 1**. The overall workflow of the Cross-species Knowledge Sharing and Preserving (CKSP) framework, which consists of data preprocessing, feature extraction, and behavioural classification. Herein, "H", "S", and "C" denote horse, sheep, and cattle, respectively.

*2.1. Data preprocessing*

Data from diverse species will initially undergo preprocessing prior to being input into the network for feature extraction. The data dimensions of different farms or institutions are normally inconsistent due to various settings (e.g., different sampling rates over identical time spans). This directly affects the feature learning capability and generalization of classification models, particularly those based on CNN-based models with fixed kernel sizes. To tackle this issue, we uniform the input dimensions across species to equal sizes. Herein, we take an example of a multi-species dataset with 2-second signal samples, sampled at 12.5 Hz for sheep, 25 Hz for cattle, and 100 Hz for horses. This results in input sizes of $1 \times 3 \times 25$, $1 \times 3 \times 50$, $1 \times 3 \times 200$, respectively. Given the balance between performance and resource consumption, a 25 Hz sampling rate is commonly adopted, as supported by recent research (Eerdekens et al., 2021; Kleanthous et al., 2022b; Riaboff et al., 2022). Therefore, we standardize these two-second inputs to a size of $1 \times 3 \times 50$, akin to 25 Hz sampled data, by exploiting the bilinear-neighbor interpolation technique (Thévenaz et al., 2000). Afterwards, these unified data are imported into the network for further feature extraction.

*2.2. Shared and preserved feature extraction*

The feature extraction phase within our proposed CKSP framework comprises of convolutional layers, batch normalization layers, max-pooling layers, global average-pooling layers, and fully connected layers, as shown in Fig. 1. Typically, different species exhibit common characteristics yet possess distinct movement patterns and divergent feature distributions. Inter-species discrepancies in movement patterns hinder conventional networks' efficiency in discerning invariant features across species, accompanied by slow convergence. Additionally, inconsistent feature distributions challenge the derivation of universal statistical measures applicable to multiple species. To tackle these challenges, we introduce a novel feature learning methodology comprised of a Shared-Preserved Convolution (SPConv) module and a Species-Specific Batch Normalization (SBN) module, as presented in Fig. 1. The SPConv module is designed with dual branches: one utilises a shared convolutional layer to distil shared universal knowledge, while the other employs species-specific convolutional layers to address the inter-species discrepancy. Following the convolutional operation, the SBN module is strategically integrated to separately fit the distributions of different species, thereby augmenting the model's adaptability to inter-species variation. The SPConv module and SBN module are detailed as follows.

*2.2.1. SPConv module*

In recent years, the practice of fine-tuning large language models has garnered growing attention due to its remarkable capability of extracting task-specific features from individual task datasets while preserving the general knowledge acquired from extensive pre-training on large data corpora (Malladi et al., 2023). Considering the substantial computational burden associated with directly fine-tuning large parameter sets, Low-Rank Adaptation (LoRA) was proposed to mitigate this issue by incorporating a low-rank parameter matrix branch, significantly reducing parameter size while effectively acquiring task-specific features (Hu et al., 2022). It enables the acquisition of domain-specific knowledge through a trainable low-rank parameter matrix, while the pre-trained full-rank parameter matrix remains fixed to preserve general knowledge.

We observe that the LoRA technique in large language model fine-tuning aligns closely with our objectives in dual aspects: the pre-trained model's function parallels our network's shared feature extraction, and the branch of low-rank parameter matrix during fine-tuning aligns with capturing species-specific information in our architecture. Inspired by this insight, we propose a Shared-Preserved Convolution (SPConv) module with dual branches: one employs a shared full-rank convolutional layer to learn shared generic features, while the other assigns individual low-rank convolutional layers to each species for the extraction of species-specific features, as shown in Fig. 1.

Let $x^s \in R^{c \times h \times w}$ represents the feature representation of species $s$ at a given layer, where $c$, $h$, and $w$ denote the channel number and spatial dimensions, respectively. We input $x^s$ into a shared full-rank convolutional layer with a $1 \times 3$ kernel to learn shared general features. Meanwhile, $x^s$ is fed into a Low-Rank Convolution (LRConv) layer to obtain personalised features of species $s$, yielding $x_o^s \in R^{c' \times h' \times w'}$. Different with a traditional $1 \times 3$ convolution operation, LRConv decomposes the original parameter tensor $W_{LR}$ of size $c' \times 3 \times 1 \times c$ into two parameter matrices: $B \in R^{(c' \times 3) \times r}$ and $A \in R^{r \times (1 \times c)}$, where $r$ denotes the low-rank value. Therefore, the LRConv operation can be formulated as:

$$x_o^s = W_{LR} x^s = R(BA) x^s, \tag{1}$$

where $BA$ denotes a matrix of dimensions $(c' \times 3) \times (1 \times c)$, and $R(BA)$ refers to reshaping matrix $BA$ into the form of $c' \times 3 \times 1 \times c$. Matrix $A$ is initialised with random Gaussian values $\mathcal{N}(0, \sigma^2)$ and $B$ is initialised as zero, implying $W_{LR} = R(BA)$ is initially set to zero during training. Notably, unlike the LoRA technique of starting with pre-trained general knowledge, our proposed method concurrently trains both shared and species-specific convolutional parameters from scratch.

*2.2.2. SBN module*

Batch normalization has been known for mitigating internal covariate shifts and enhancing feature discriminability while accelerating the learning process (Ioffe & Szegedy, 2015). It typically operates within a batch, normalizing the values across each channel by adjusting them to have zero mean and unit variance across the entire batch. Following this normalization, an affine transformation, parameterized by trainable parameters $[\gamma, \beta]$, is imposed on the normalised feature maps. During the training process, the BN layer is able to capture global statistics measures – a moving mean and a moving variance – which are later fixed and utilised to normalise features in the testing phase, ensuring stable performance.

The efficacy of BN operations in prevailing studies largely relies on the assumption that training data originates from the same species and follows a uniform distribution. However, the data collected from diverse animal species in our study exhibit distinct movement patterns, thereby posing a challenge to acquire universal global statistical measures viable across various species. As illustrated in Fig. 2, we visualise the species-specific global statistics obtained from the BN layers of networks separately trained on data from each species (e.g., horse, sheep, and cattle). It can be observed that these statistics display notable discrepancies between different species, particularly in the deeper layers that contain more discriminative and semantically rich features. Directly aggregating these data with inherent statistical variations for joint training inevitably imposes difficulties on the network to learn generic features and impedes model convergence. In addition, the shared statistical parameters, as learned, inadequately mirror the feature distribution of individual species, thereby undermining the classification performance of models during the testing phase across diverse species.

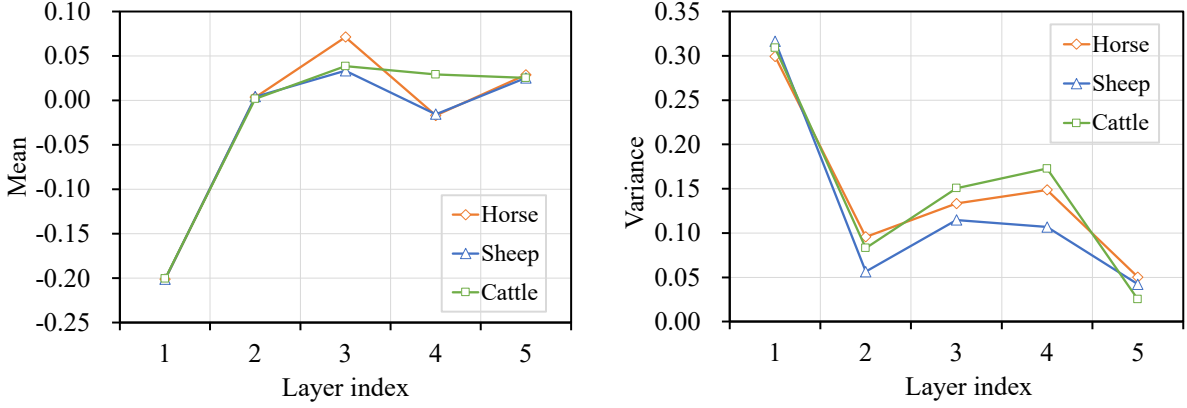

**Fig. 2**. The moving mean and variance derived from the classification model trained exclusively on single-species data. These values represent the layer-wise averages of the batch normalization statistics.

To address the aforementioned issue, motivated by the efficacy of employing separate BN layers in handling data heterogeneity from diverse origins (X. Li et al., 2021; Liu et al., 2020), we design a Species-specific Batch Normalization (SBN) module. The SBN module assigns an individual BN layer for each species and is incorporated subsequent to each convolutional or fully connected layer. Specifically, the SBN module sets each animal species $s$ with respective trainable variables $[\gamma^s, \beta^s]$. Let $x_b^s \in [x_1^s, ..., x_B^s]$ represents the feature map of the $b$-th sample within a batch of species $s$ for a certain channel at a given layer, the corresponding output $y_b^s$ can be formulated as:

$$y_b^s = \gamma^s \cdot \tilde{x}_b^s + \beta^s, \qquad \text{with } \tilde{x}_b^s = \frac{x_b^s - E[x_b^s]}{\sqrt{Var[x_b^s] + \epsilon}}. \tag{2}$$

The learned global statistics $[\gamma^s, \beta^s]$, i.e., the moving mean and moving variance, are then utilised to normalise features extracted from test data pertaining to species $s$. Through individual feature normalization, the model is capable of learning precise statistics tailored to each species, which in turn accelerates convergence and boosts the model's classification capabilities. Herein, we substitute the standard shared convolutional layer and BN layer with our proposed SPConv and SBN modules in all layers except the initial one. This is based on the findings that shallower layers typically concentrate on generic feature extraction (Lang et al., 2022), and the initial layer exhibits less pronounced differences in feature distributions among species (Fig. 2).

*2.3. Behavioural classification*

Considering the behavioural disparities among different animal species, to enhance the applicability of models across various species, species-specific classifiers are appended following the feature extraction stage. As presented in Fig. 1, for each animal species, a separate classifier comprised of a single fully connected layer is adopted. Herein, we apply class-balanced focal loss as the loss function, which has been validated in addressing class imbalance problems (Cui et al., 2019; Mao et al., 2021). To guarantee the model remains unbiased towards any one species and can concurrently learn knowledge from distinct species, we evenly distribute data from various species within each batch during the training process. Hence, the overall loss function can be formulated as the average of losses across all $S$ species:

$$\mathcal{L} = \frac{\sum_{s=1}^{S} \mathcal{L}^s}{S}, \qquad (3)$$

where $\mathcal{L}^s$ denotes the class-balanced focal loss of species $s$. Utilising the loss in Eq. (3), our proposed CKSP framework can enhance the classification performance for individual species and flexibly accommodate diverse behavioural categories across different species.

## 3. Datasets and experimental setup

*3.1. Datasets*

Our proposed CKSP is evaluated utilising three publicly available datasets collected from horses (Kamminga et al., 2019a), sheep (Kleanthous et al., 2022a) and cattle (C. Li et al., 2021), respectively. The specifics are summarized in the following Table 1.

**Horse dataset.** The horse dataset encompasses 87,621 two-second samples, acquired from six horses using neck-attached inertial measurement units with a sampling rate of 100 Hz. Based on existing studies on horse behaviour recognition (Kamminga et al., 2019; Mao et al., 2023b), we consider five extensively labelled activities, including grazing, galloping, standing, trotting, and walking. Amongst, the triaxial accelerometer measurements were employed, resulting in a tensor shape of $1 \times 3 \times 200$ for each two-second sample.

*Sheep dataset*. The sheep dataset consists of 149,725 two-second motion data collected from nine sheep using neck-attached accelerometers with a sampling rate of 12.5 Hz. Five activities, including grazing, walking, scratching, standing, and resting, are contained and merged into three main unified behaviours, i.e., grazing, active (including walking and scratching), and inactive (including standing and resting). The triaxial accelerometer data construct a tensor of dimensions $1 \times 3 \times 25$ for each sample.

*Cattle dataset*. The cattle dataset is collected from six different Japanese black beef cows using neck-attached accelerometers with a sampling rate of 25 Hz. Based on existing studies on cattle behaviour recognition (Arablouei et al., 2023a; C. Li et al., 2021; Minati et al., 2023), we consider five frequent cattle behaviours, including grazing, ruminating, resting, moving, and salting. The dataset contains a total of 10,429 two-second data samples, with each sample comprising triaxial accelerometer data structured as a tensor of $1 \times 3 \times 50$.

**Table 1.** Illustration of datasets collected from cattles, horses, and sheep.

| Dataset | Object number | Sampling rate | Activity | Data Number | Reference |
|---|---|---|---|---|---|
| Horse | 6 | 100 | Grazing, galloping, standing, trotting, and walking | 87,621 | (Kamminga, Janßen, et al., 2019) |
| Sheep | 9 | 12.5 | Grazing, active (walking and scratching), and inactive (standing and resting) | 149,725 | (Kleanthous, Hussain, Khan, Sneddon, & Liatsis, 2022) |
| Cattle | 6 | 25 | Grazing, moving, resting, ruminating, and salting | 10,429 | (C. Li et al., 2021) |

*3.2. Experimental setup*

To ensure the model remains unbiased towards any individual species and demonstrate our approach's effectiveness under data limitation scenarios, we equalise the training data size across all species by downsampling to match them to the quantity of the species with the smallest number. These sampled data are subsequently combined to train our proposed model, where the sample number within each batch should be uniform across different species during training. Precision, recall, F1-score, and accuracy are used as evaluation metrics to gauge the overall performance of the classification network. To validate the generalisation ability of our approach, we perform the stratified 5-fold cross-validation method where samples of three, one, and one folds are separated as the training, validation, and testing datasets, respectively.

During the training, $L_2$ regularization with a weight decay of 0.06 is applied to the loss function to mitigate overfitting. An Adam optimiser initiates training with a learning rate of $1 \times 10^{-4}$, which is reduced by a factor of 0.1 every 20 epochs. The training runs for 100 epochs with a batch size of 256. The model achieving the highest validation accuracy is saved and evaluated on the test set for verification. To evaluate the proposed CKSP, we contrast it directly with a baseline model (Single-Net), which is trained exclusively on data from individual species. All tests are performed utilising the PyTorch platform on an NVIDIA GeForce RTX 2080 graphics processing unit. The source code will be available at https://github.com/Max-1234-hub/CKSP.

## 4. Results and discussion

Overall, the experimental results highlight the significant superiority of our CKSP approach over the Single-Net model trained solely on individual species data. Ablation studies confirm the effectiveness of the SPConv and SBN components in enhancing classification performance. Furthermore, the recognition analysis illuminates the predictive advantages of CKSP in data-constrained settings. This section concludes with suggestions for future research directions.

*4.1. Performance comparisons with the baseline method*

To assess the performance of our proposed method, we compare CKSP against Single-Net and present the results in Fig. 3, with both models trained on datasets from three different species. The results reveal that our proposed CKSP exhibits promising performance, achieving accuracies of 96.44%, 92.89%, 90.01% on the horse, sheep, and cattle datasets respectively, accompanied by F1-scores of 96.02%, 86.79%, and 88.40%, precision values of 95.07%, 87.39%, and 85.76%, and recall values of 97.03%, 86.59%, and 91.60% on the respective datasets. Obviously, the CKSP outperforms the Single-Net in terms of all evaluation metrics, with increments of 6.04%, 2.06%, and 3.66% in accuracy, 10.33%, 3.67%, and 7.90% in F1-score, 12.46%, 3.87%, and 8.96% in precision, and 6.24%, 3.66%, and 4.03%

in recall for the horse, sheep, and cattle datasets, respectively. This demonstrates the promising capabilities of our method in leveraging multi-species data to augment classification performance.

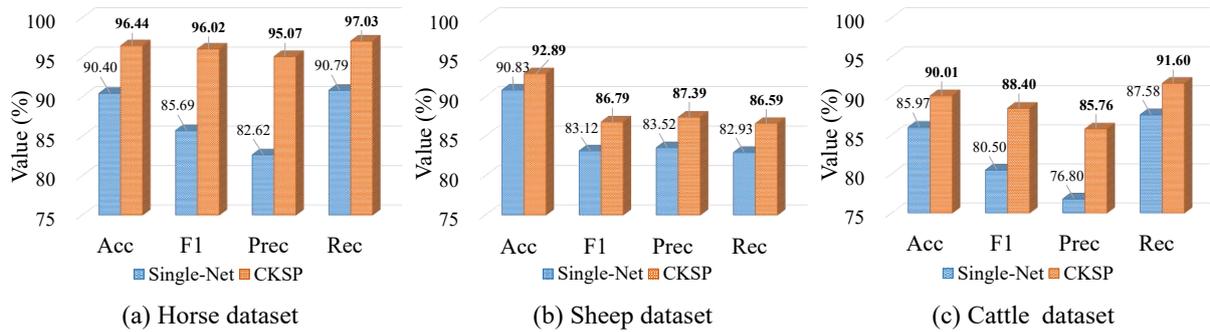

**Fig. 3**. Comparison results of our proposed CKSP with the Single-Net on the horse (a), sheep (b), and cattle (c) datasets. Herein, "Acc", "F1", "Prec", and "Rec" denote accuracy, F1-score, precision, and recall, respectively.

Figure 4 illustrates the recall confusion matrices for Single-Net and our CKSP method, revealing the latter's notable potential to improve classification accuracy across diverse activity categories. Specifically, recall represents the percentage of correctly classified samples. Compared to the Single-Net (Fig. 4a), the proposed CKSP method (Fig. 4b) significantly elevates recall values for nearly all activities, demonstrating varying degrees of increment. Particularly, our method elevates the recognition accuracy for various horse behaviours above 95%, and for cattle, around 90% or even higher. Even though the classification accuracy for sheep activities including grazing and active is undesirable, our method has managed to enhance these metrics by 6.68% and 4.18%, respectively. It can be observed that grazing and active behaviours in sheep are prone to misclassification with one another, aligning with findings reported in Kleanthous et al.' (2022a) study on sheep activity recognition. This might be accounted for by the similarity in movements exhibited during grazing and active behaviours in sheep in some cases. Thus, exploring potential solutions to alleviate the impact of activity similarities in classification performance is a venture deserving of attention in forthcoming studies.

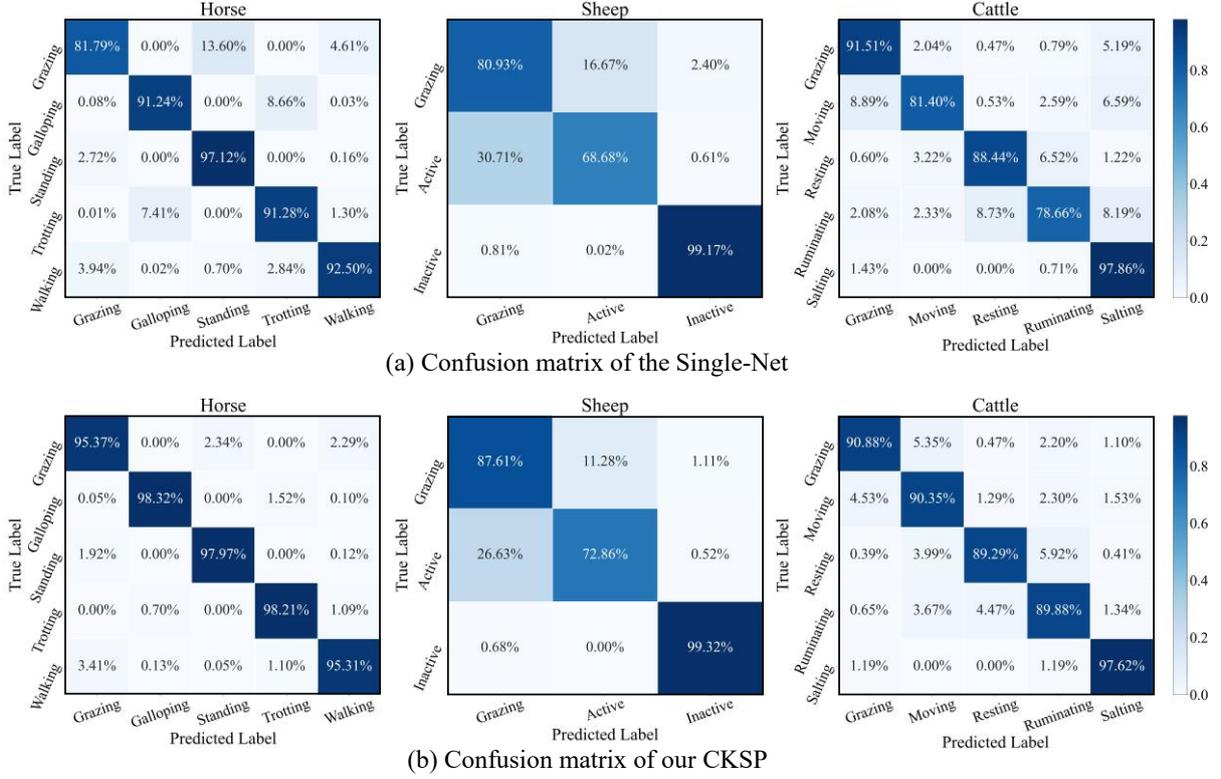

Fig. 4. Confusion matrices of the Single-Net (a) and our CKSP (b) on the horse, sheep, and cattle datasets.

### 4.2. Ablation studies

#### 4.2.1. Evaluation of SPConv and SBN modules

To thoroughly probe the impacts of the SPConv and SBN modules, we perform experiments on the proposed CKSP framework with and without SPConv and/or SBN modules. The results obtained on the three datasets are given in Table 2. We can see that our method without both modules, i.e., directly sharing feature extraction parameters across distinct animal species, yields inferior results. Conversely, the integration of the SBN module enables our method to obtain premium performance with different degrees of improvement, implying the critical importance of fitting species-specific feature distributions for different species. Notably, including the SPConv module alongside the SBN module yields additional enhancements to classification performance, evidenced by accuracy increments of 8.67%, 2.02%, 11.64%; F1-score gains of 13.69%, 3.66%, 19.79%; precision boosts of 15.84%, 3.93%, 17.26%; and recall increases of 7.13%, 3.74%, 11.93% for the horse, sheep, and cattle datasets, respectively. This confirms our earlier assertion that a combination of shared and personalised learning parameters is necessary, given the coexistence of both generic and species-specific behavioural patterns across different species. However, the exclusive use of the SPConv module without the SBN module results in diminished performance, potentially due to the conflicting nature of learning species-specific parameters while still assuming a shared feature distribution.

**Table 2.** Ablation results of the proposed CSKP framework, assessing its performance with and without SPConv and/or SBN modules.

| Configurations | | Horse | | | | Sheep | | | | Cattle | | | |
|---|---|---|---|---|---|---|---|---|---|---|---|---|---|
| SPConv | SBN | Acc[#] (%) | F1 (%) | Prec (%) | Rec (%) | Acc (%) | F1 (%) | Prec (%) | Rec (%) | Acc (%) | F1 (%) | Prec (%) | Rec (%) |
| | | 61.16 | 67.59 | 75.91 | 70.44 | 73.51 | 52.92 | 70.10 | 50.15 | 71.52 | 60.53 | 68.88 | 69.79 |
| √ | | 76.97 | 62.32 | 63.92 | 67.85 | 60.60 | 49.23 | 61.68 | 56.51 | 63.77 | 41.19 | 48.15 | 46.53 |
| | √ | 87.77 | 82.33 | 79.23 | 89.90 | 90.87 | 83.13 | 83.46 | 82.85 | 78.37 | 68.61 | 68.50 | 79.67 |
| √ | √ | **96.44**[*] | **96.02** | **95.07** | **97.03** | **92.89** | **86.79** | **87.39** | **86.59** | **90.01** | **88.40** | **85.76** | **91.60** |

[#] Acc: accuracy; F1: F1-score; Prec: precision; Rec: recall.

[*] The best result for each metric is highlighted in bold.

Figure 5 displays the training and validation accuracy curves across three distinct species for models configured without and with the combined utilisation of SPConv and SBN modules. We can find that our CKSP method, when equipped with both SPConv and SBN modules, converges faster and exhibits a smoother training trajectory compared to the configuration lacking these modules. This reflects that CKSP could effectively mitigate data heterogeneity and expedite convergence by extracting robust representations from multi-species data.

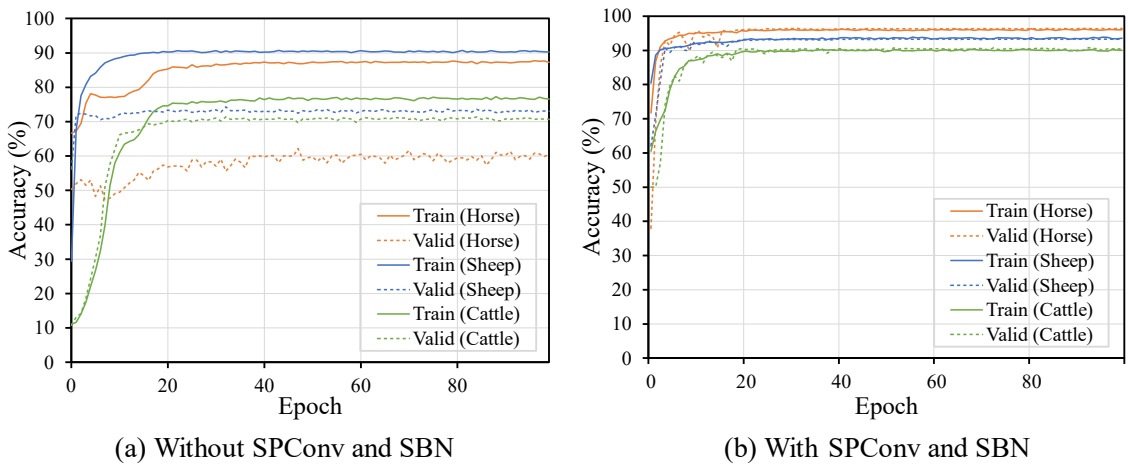

**Fig. 5**. The training and validation accuracy over three different animal species under the CKSP without (a) and with (b) both the SPConv and SBN modules.

*4.2.2. Analysis of the SPConv module*

***Analysis of the LRConv operation.*** The LRConv operation adapts the species-specific convolution layer by introducing a low-rank decomposition, effectively decreasing the parameters within a certain scope. To gain insights into the benefits of LRConv operation, we contrast the performance of our CKSP framework equipped with LRConv layers against a variant using full-rank convolution (FRConv) layers, as detailed in Table 3. It is obvious that the CKSP employing LRConv layers demonstrates superior performance compared to the version utilising FRConv layers, regardless of the varied values of $r$ (2~16). This evidences the efficacy of our implemented LRConv layers in enhancing performance. Moreover, the total parameter count within the species-specific branches adopting LRConv layers is smaller than those using FRConv layers, highlighting the additional advantage of employing LRConv layers in terms of efficiency.

***Analysis of the hyper-parameter $r$.*** The hyper-parameter $r$ in the SPConv module denotes the low-rank value. Herein, we analyse the performance of our method with varying $r$ values (i.e., 2, 4, 8, 12, and 16) through experiments, and the findings are summarized in Table 3. When $r$ is set to 12, the CKSP attains the highest values across all evaluation metrics for horse and cattle behaviour classification and demonstrates favourable performance in classifying sheep behaviours. It underscores the potential advantage of judiciously selecting the value of $r$ for enhancing the overall classification performance.

**Table 3.** Experimental results comparing CKSP integrated with full-rank convolution (FRConv) layer and low-rank convolution (LRConv) layers across different $r$ values.

| SPConv | | Horse | | | | Sheep | | | | Cattle | | | |
|---|---|---|---|---|---|---|---|---|---|---|---|---|---|
| | | Acc[#] (%) | F1 (%) | Prec (%) | Rec (%) | Acc (%) | F1 (%) | Prec (%) | Rec (%) | Acc (%) | F1 (%) | Prec (%) | Rec (%) |
| FRConv | | 93.07 | 89.92 | 87.16 | 94.03 | 90.72 | 82.71 | 83.07 | 82.61 | 84.76 | 78.40 | 75.17 | 85.48 |
| LRConv | $r$=2 | 94.26 | 92.35 | 90.09 | 95.23 | 91.79 | 84.73 | 85.22 | 84.55 | 87.70 | 84.46 | 80.94 | 89.38 |
| | $r$=4 | 95.40 | 94.16 | 92.41 | 96.21 | 92.44 | 86.05 | 86.40 | 85.96 | 88.91 | 86.28 | 82.92 | 90.61 |
| | $r$=8 | 95.79 | 95.20 | 93.83 | 96.71 | 92.80 | 86.61 | 87.24 | 86.40 | 89.26 | 87.87 | 85.05 | 91.43 |
| | $r$=12 | **96.44**[*] | **96.02** | **95.07** | **97.03** | 92.89 | 86.79 | 87.39 | 86.59 | **90.01** | **88.40** | **85.76** | **91.60** |
| | $r$=16 | 95.23 | 95.08 | 93.77 | 96.59 | **92.96** | **86.90** | **87.45** | **86.68** | 89.04 | 87.41 | 84.38 | 91.22 |

[#] Acc: accuracy; F1: F1-score; Prec: precision; Rec: recall.

[*] The best result for each metric is highlighted in bold.

*4.3. Robustness against variations in dataset size*

The proposed CKSP method leverages multi-species datasets to establish a universal AAR framework. This strategy facilitates the capture of a broader spectrum of movement patterns across diverse species, thereby providing a potential prospect to alleviate the poor performance resulting from insufficient sample sizes of a single species. To validate the classification ability of our CKSP approach under the context of data limitation, we present in Fig. 6 the comparative classification performance between the Single-Net and CKSP over varying percentages (i.e., 75%, 50%, 25%, and 10%) of the original dataset. The CKSP exhibits remarkable stability in classifying horse and sheep behaviours, and the improvement margin it gains over the Single-Net increases as the dataset size decreases. The findings indicate the robustness of our method, and it can effectively benefit from the diversity of multi-species datasets, particularly in scenarios characterized by data scarcity. Consistently, our method's performance in cattle classification echoes the trends observed in the preceding species (horse and sheep) as the dataset shrinks from 100% to 25%. Despite a conspicuous drop in data percentage of 10%, it persistently surpasses the Single-Net's performance. This phenomenon might be attributed to certain behaviours unique to cattle, i.e., ruminating and salting; when the sample size decreases to a certain threshold, even aggregating data from diverse species does not sufficiently augment the diversity of these specific behaviours, thereby imposing limitations on performance improvement.

*4.4. Limitations and implications*

The proposed CKSP approach can be applicable to diverse species while mitigating the challenge of data limitation by learning cross-species features. Nevertheless, the efficacy of enhancing diversity through aggregating multi-species datasets typically hinges on the prerequisite that two or more species have comparable behavioural categories. To this point, we will establish a universal and standardized dictionary of behaviours through extensive research and field studies, with each behaviour being linked to animal health and well-being. This dictionary will serve as a reference for future researchers, who are encouraged to collect data based on their areas of interest within this dictionary and, where possible, strive for data openness. This collaborative endeavour paves the way for developing a large AAR model grounded in a universal database, laying a robust foundation for upcoming advancements.

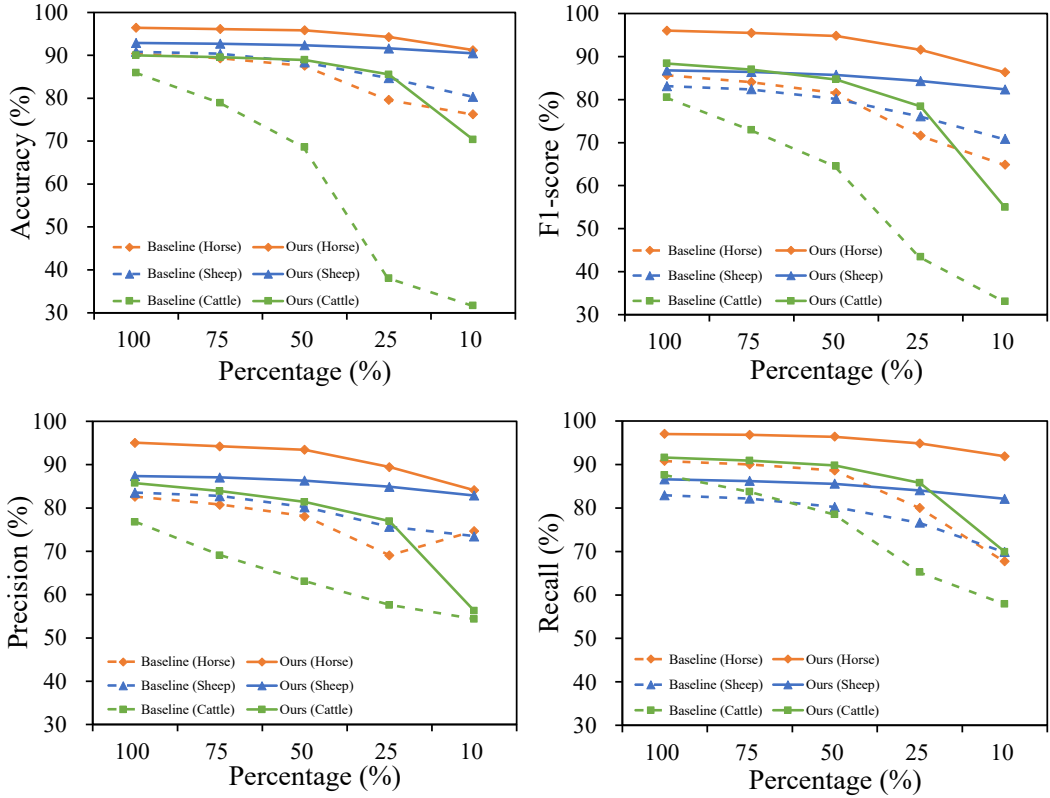

**Fig. 6**. Classification performance of the Single-Net and our CKSP approach over varying data sizes.

Our current challenge resides in discerning inter-activity similarity, where distinct animal behaviours exhibit similar characteristics or movement patterns (Mao et al., 2023a). This hinders deep learning models from extracting discriminative features that uniquely identify activities, leading to confusion in classifying them (Chen et al., 2021), as evidenced by the challenge of differentiating sheep grazing from active behaviour in this study. Hence, our next step to refine classification accuracy for similar activities involves exploring viable approaches such as fine-grained activity recognition (De et al., 2015), context-aware modelling (Yurur et al., 2014), and integrating multiple wearable sensor types (Halachmi et al., 2019).

## 5. Conclusions

This study develops a universal AAR framework named CKSP involving an SPConv module and an SBN module, based on sensor data across diverse animal species. The CKSP is applicable to diverse species with distinct interested behaviours while mitigating the challenge of data limitation by learning cross-species features. Considering the coexistence of both similarities and differences of behaviours among different species, the SPConv module assigns individual low-rank convolutional layers to each species for extracting species-specific features, while employing a shared full-rank convolutional layer to learn generic features. Given that different species exhibit distinct data distributions, the SBN module allocates a separate BN layer to each species, independently adapting to the unique distribution characteristics of each, thereby enhancing normalization efficacy across diverse species. The experimental outcomes reveal that our CKSP method surpasses Single-Net, trained exclusively on species-specific data. Ablation studies underscore the efficacy and importance of each component in our approach, emphasizing their contribution to overcoming challenges posed by limited sample sizes of individual species. In short, this work opens up a potential pathway for developing a large-scale AAR model, thereby advancing the field of precision livestock farming.

**CrediT Authorship Contribution Statement**

**Axiu Mao:** Conceptualization; Methodology; Software; Validation; Formal analysis; Resources; Supervision; Project administration; Funding acquisition; Writing - original draft. **Meilu Zhu:** Conceptualization; Methodology; Investigation; Validation; Writing - review & editing. **Zhaojin Guo:** Data curation. **Zheng He:** Visualization. **Tomas Norton:** Writing - review & editing. **Kai Liu:** Conceptualization; Writing - review & editing.

**Data availability statement**

The data used in the current study are open-source data, which can be accessed at: https://data.4tu.nl/articles/_/12687551/1 (Horse), https://zenodo.org/record/5849025#.ZE-y_3ZByHu (Cattle), https://github.com/nkleanthous2015/Sheep_activity_Data (Sheep).

**Acknowledgements**

This work was supported by the "new research initiatives" (grant numbers: KYS085624257) at Hangzhou Dianzi University.

**Statement on the Use of Generative AI and AI assisted technologies in the writing process**

No generative AI or AI-assisted technologies were used during the preparation of this work.